\def\year{2016}
\documentclass[letterpaper]{article}

\usepackage{aaai16}
\usepackage{times}
\usepackage{helvet}
\usepackage{courier}
\usepackage{times}
\usepackage{url}
\usepackage{latexsym}
\usepackage{amssymb}
\usepackage{amsmath}
\usepackage{graphicx}
\usepackage{multirow}
\usepackage{tabularx}
\usepackage{lineno}
\usepackage[font=small,labelfont=bf]{caption}
\usepackage{booktabs}
\newcolumntype{Y}{>{\centering\arraybackslash}X}
\DeclareMathOperator{\Tr}{Tr}
 
\begin{document}

%
\title{Character-Aware Neural Language Models}
\author{
Yoon Kim$^\dag$ \\ \\
\hspace{48mm} $^\dag$School of Engineering and Applied Sciences \\
\hspace{48mm} Harvard University \\
\hspace{48mm} \texttt{\{yoonkim,srush\}@seas.harvard.edu}
\And
Yacine Jernite$^*$\\ 
\And
David Sontag$^*$ \\ \\
\hspace{50mm}$^*$Courant Institute of Mathematical Sciences \\
\hspace{50mm}New York University \\
\hspace{50mm}\texttt{\{jernite,dsontag\}@cs.nyu.edu}
\And
Alexander M. Rush$^\dag$ \\ 
}
\maketitle
\begin{abstract}
\begin{quote}
We describe a simple neural language model that relies only on character-level inputs. 
Predictions are still made at the word-level. Our model employs a convolutional neural 
network (CNN) and a highway network over characters, whose output is given to a 
long short-term memory (LSTM) recurrent neural network language model (RNN-LM). 
On the English Penn Treebank the model is on par with the existing state-of-the-art despite 
having $60\%$ fewer parameters. On languages with rich morphology (Arabic, Czech, French, German, 
Spanish, Russian), the model outperforms word-level/morpheme-level LSTM baselines, again with fewer parameters. 
The results suggest that on many languages, character inputs are sufficient for language modeling. Analysis of
word representations obtained from the character composition part of the model reveals that 
the model is able to encode, from characters only, both semantic and orthographic information.
\end{quote}
\end{abstract}

\section{Introduction}

Language modeling is a fundamental task in artificial intelligence and natural language processing (NLP), with applications in speech recognition,
text generation, and machine translation. A language model is formalized as a probability distribution 
over a sequence of strings (words), and traditional methods usually involve making an $n$-th order Markov assumption and estimating
$n$-gram probabilities via counting and subsequent smoothing \cite{Chen1998}. The count-based models are simple to train, but
 probabilities of rare $n$-grams can be poorly estimated due to data sparsity (despite smoothing techniques).

Neural Language Models (NLM) address the $n$-gram data sparsity issue through parameterization of words as vectors 
(word embeddings) and using them as inputs to a neural network \cite{Bengio2003,Mikolov2010}.
The parameters are learned as part of the training process. Word embeddings obtained through 
NLMs exhibit the property whereby semantically close words are likewise close in the induced vector space
(as is the case with non-neural techniques such as Latent Semantic Analysis \cite{Deerwester1990}). 

While NLMs have been shown to outperform count-based $n$-gram language models \cite{Mikolov2011}, they are 
blind to subword information (e.g. morphemes). For example, they do not know, a priori, that \emph{eventful}, \emph{eventfully},
\emph{uneventful}, and \emph{uneventfully} should have structurally related embeddings in the
 vector space. Embeddings of rare words can thus be poorly estimated, 
leading to high perplexities for rare words (and words surrounding them). This is especially problematic in 
morphologically rich languages  with long-tailed frequency distributions or domains with
dynamic vocabularies (e.g. social media).

In this work, we propose a language model that leverages subword information through a character-level convolutional neural network
(CNN), whose output is used as an input to a recurrent neural network language model (RNN-LM). Unlike previous
works that utilize subword information via morphemes \cite{Botha2014,Luong2013}, our model does not require morphological tagging
as a pre-processing step. And, unlike the recent line of work which combines input word embeddings with features from a character-level model
 \cite{Santos2014a,Santos2015}, our model does not utilize word embeddings at all in the input layer. Given that most of the parameters in
NLMs are from the word embeddings, the proposed model has significantly fewer parameters than previous NLMs, making it attractive for 
applications where model size may be an issue (e.g. cell phones). 

To summarize, our contributions are as follows:
\begin{itemize}
\item on  English, we achieve results on par with the existing state-of-the-art on the Penn Treebank (PTB),
despite having approximately $60\%$ fewer parameters, and
\item on morphologically rich languages (Arabic, Czech, French, German, Spanish, and Russian), our model outperforms various baselines (Kneser-Ney, 
word-level/morpheme-level LSTM), again with fewer parameters.
\end{itemize}
We have released all the code for the models described in this paper.\footnote{\url{https://github.com/yoonkim/lstm-char-cnn}}

\section{Model}
\begin{figure}[!t]
\center
\includegraphics[scale=0.50]{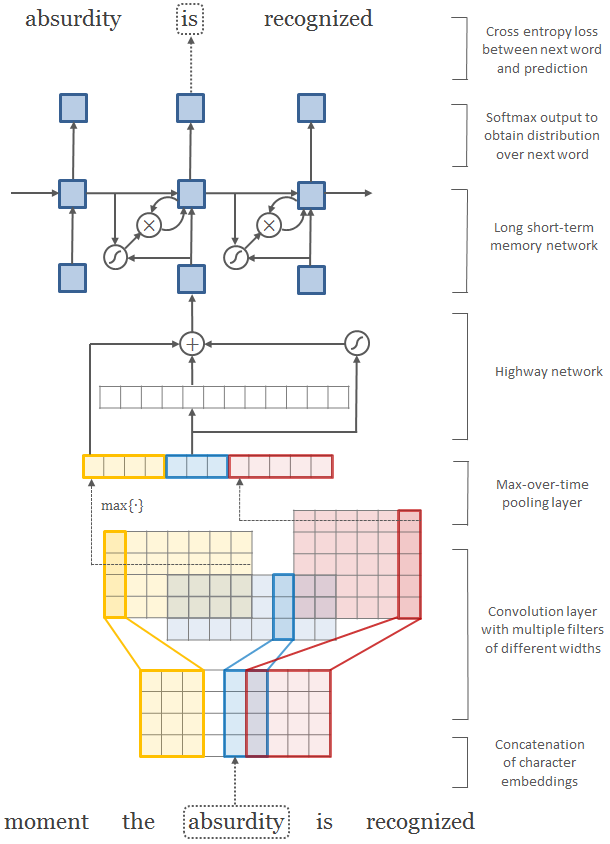}
\caption{Architecture of our language model applied to an example sentence. Best viewed in color. Here the model takes {\em absurdity}
 as the current input and combines it with the history (as represented by the hidden state) to predict the next word, {\em is}. 
First layer performs a lookup of character embeddings (of dimension four) and stacks them to form the matrix $\mathbf{C}^k$. 
Then convolution operations are applied between $\mathbf{C}^k$ and multiple filter matrices. Note that in the above example we have 
twelve filters---three filters of width two (blue), four filters of width three (yellow), and five filters of width four (red). 
A max-over-time pooling operation is applied to obtain a fixed-dimensional representation of the word, which is given to the highway network.
The highway network's output is used as the input to a multi-layer LSTM. Finally, an affine transformation followed by a softmax 
is applied over the hidden representation of the LSTM to obtain the distribution over the next word. Cross entropy loss between the
(predicted) distribution over next word and the actual next word is minimized.
Element-wise addition, multiplication, and sigmoid operators are 
depicted in circles, and affine transformations (plus nonlinearities where appropriate) are represented by solid arrows.}
\label{fig:network}
\end{figure}

The architecture of our model, shown in Figure~\ref{fig:network}, is straightforward. Whereas a conventional NLM takes word embeddings as inputs, our
model instead takes the output from a single-layer character-level convolutional neural network with max-over-time pooling.
 
For notation, we denote vectors with bold lower-case (e.g. $\mathbf{x}_t,\mathbf{b}$), matrices with bold upper-case (e.g. $\mathbf{W}, \mathbf{U}^o$), 
scalars with italic lower-case (e.g. $x,b$), and sets with cursive upper-case (e.g. $\mathcal{V}, \mathcal{C}$) letters. 
For notational convenience we assume that words and characters have already been converted into indices.

\subsection{Recurrent Neural Network}

A recurrent neural network (RNN) is a type of neural network architecture particularly suited for modeling sequential phenomena.
At each time step $t$, an RNN takes the input vector $\mathbf{x}_t \in \mathbb{R}^n$ and the hidden state vector  
$\mathbf{h}_{t-1} \in \mathbb{R}^m$ and produces the next hidden state $\mathbf{h}_t$ by applying the following recursive operation:
\begin{equation}
\mathbf{h}_t = f(\mathbf{W} \mathbf{x}_t + \mathbf{U} \mathbf{h}_{t-1} + \mathbf{b})
\end{equation}
Here $\mathbf{W} \in \mathbb{R}^{m \times n}, \mathbf{U} \in \mathbb{R}^{m \times m}, \mathbf{b} \in \mathbb{R}^{m}$ are parameters of an 
affine transformation and $f$ is an element-wise nonlinearity. In theory the RNN can summarize all historical information
up to time $t$ with the hidden state $\mathbf{h}_t$. In practice however, learning long-range dependencies with a vanilla RNN is difficult due
to vanishing/exploding gradients \cite{Bengio1994}, which occurs as a result of the Jacobian's multiplicativity with respect to time.

Long short-term memory (LSTM) \cite{Hochreiter1997} addresses the problem of learning long range dependencies by augmenting the RNN
with a memory cell vector $\mathbf{c}_t \in \mathbb{R}^n$ at each time step. Concretely, one step of
an LSTM takes as input $\mathbf{x}_t, \mathbf{h}_{t-1}, \mathbf{c}_{t-1}$ and produces $\mathbf{h}_t$, $\mathbf{c}_t$ via the
following intermediate calculations:
\begin{equation}
\begin{split}
\mathbf{i}_t &= \sigma (\mathbf{W}^i \mathbf{x}_t + \mathbf{U}^i \mathbf{h}_{t-1} + \mathbf{b}^i) \\
\mathbf{f}_t &= \sigma (\mathbf{W}^f \mathbf{x}_t + \mathbf{U}^f \mathbf{h}_{t-1} + \mathbf{b}^f) \\
\mathbf{o}_t &= \sigma (\mathbf{W}^o \mathbf{x}_t + \mathbf{U}^o \mathbf{h}_{t-1} + \mathbf{b}^o) \\
\mathbf{g}_t &= \mbox{tanh} (\mathbf{W}^g \mathbf{x}_t + \mathbf{U}^g \mathbf{h}_{t-1} + \mathbf{b}^g) \\
\mathbf{c}_t &= \mathbf{f}_t \odot \mathbf{c}_{t-1} + \mathbf{i}_t \odot \mathbf{g}_t \\
\mathbf{h}_t &= \mathbf{o}_t \odot \mbox{tanh} (\mathbf{c}_t)
\end{split}
\end{equation}
Here $\sigma(\cdot)$ and $\mbox{tanh}(\cdot)$ are the element-wise sigmoid and hyperbolic tangent functions, $\odot$ is the element-wise multiplication operator,
and $\mathbf{i}_t$,  $\mathbf{f}_t$,  $\mathbf{o}_t$ are referred to as {\em input}, {\em forget}, and {\em output} gates. At $t=1$, $\mathbf{h}_0$ and $\mathbf{c}_0$ 
are initialized to zero vectors. Parameters of the LSTM are $\mathbf{W}^j, \mathbf{U}^j, \mathbf{b}^j$ for $j \in \{i, f, o, g\}$. 

Memory cells in the LSTM are additive with respect to time, alleviating the gradient vanishing problem. Gradient exploding is still an issue,
 though in practice simple optimization strategies (such as gradient clipping) work well. LSTMs have been shown to outperform 
vanilla RNNs on many tasks, including on language modeling \cite{Sundermeyer2012}. It is easy to extend the RNN/LSTM to two (or more) 
layers by having another network whose input at $t$ is $\mathbf{h}_t$ (from the first network). Indeed, 
having multiple layers is often crucial for obtaining competitive performance on various tasks \cite{Pascanu2013}.

\subsection{Recurrent Neural Network Language Model}

Let $\mathcal{V}$ be the fixed size vocabulary of words. A language model specifies a distribution over $w_{t+1}$ (whose support is $\mathcal{V}$) 
given the historical sequence $w_{1:t} = [w_1, \dots , w_t]$. A recurrent neural network language model (RNN-LM) does this by applying an 
affine transformation to the hidden layer followed by a softmax:
\begin{equation}
\mbox{Pr}(w_{t+1} = j|w_{1:t}) = \frac{\mbox{exp}(\mathbf{h}_t \cdot \mathbf{p}^j + q^j)}{\sum_{j' \in \mathcal{V}} 
\mbox{exp}(\mathbf{h}_t \cdot \mathbf{p}^{j'} + q^{j'})}
\end{equation}
where $\mathbf{p}^j$ is the $j$-th column of $\mathbf{P} \in \mathbb{R}^{m \times |\mathcal{V}|}$  (also referred to as the 
{\em output embedding}),\footnote{In our work, predictions are at the word-level, and hence we still utilize word embeddings in the output layer.} 
and $q^j$ is a bias term. Similarly, for a conventional RNN-LM which usually takes words as inputs,  if $w_t = k$, then the input to the RNN-LM at $t$ 
is the {\em input embedding}  $\mathbf{x}^k$, the $k$-th column of the embedding matrix $\mathbf{X} \in \mathbb{R}^{n \times |\mathcal{V}|}$.
Our model simply replaces the input embeddings $\mathbf{X}$ with the output from a character-level convolutional neural network, to be described below.

If we denote $w_{1:T} = [w_1, \cdots, w_T]$ to be the sequence of words in the training corpus, training involves minimizing the 
negative log-likelihood ($NLL$) of the sequence
\begin{equation}
NLL = -\sum_{t=1}^T \mbox{log } \mbox{Pr}(w_{t} | w_{1:t-1})
\end{equation}
which is typically done by truncated backpropagation through time \cite{Werbos1990,Graves2013}. 

\subsection{Character-level Convolutional Neural Network}

In our model, the input at time $t$ is an output from a character-level convolutional neural network (CharCNN), which we describe in this section.  
CNNs \cite{LeCun1989} have achieved state-of-the-art results on computer vision \cite{Krizhevsky2012} and have also been shown to be effective 
for various NLP tasks \cite{Collobert2011}. Architectures employed for NLP applications differ in that they typically involve temporal rather 
than spatial convolutions.

Let $\mathcal{C}$ be the vocabulary of characters, $d$ be the dimensionality of character embeddings,\footnote{Given that $|\mathcal{C}|$ 
is usually small, some authors work with one-hot representations of characters. However we found that using lower dimensional representations of 
characters (i.e. $d < |\mathcal{C}|$) performed slightly better.} and $\mathbf{Q} \in \mathbb{R}^{d \times |\mathcal{C}|}$  be the matrix character embeddings.
Suppose that word $k \in \mathcal{V}$ is made up of a sequence of characters $[c_1, \dots, c_{l}]$, where $l$ is the length of word $k$.
Then the character-level representation of $k$ is given by the matrix $\mathbf{C}^k \in \mathbb{R}^{d \times l}$, where the $j$-th column corresponds 
to the character embedding for $c_j$ (i.e. the $c_j$-th column of $\mathbf{Q}$).\footnote{Two technical details warrant mention here: (1) we append 
start-of-word and end-of-word characters to each word to better represent prefixes and suffixes and hence $\mathbf{C}^k$ actually has $l + 2$ columns;
 (2) for batch processing, we zero-pad $\mathbf{C}^k$ so that the number of columns is constant (equal to the max
word length) for all words in $\mathcal{V}$.}

We apply a narrow convolution between $\mathbf{C}^k$ and a {\em filter} (or {\em kernel}) $\mathbf{H} \in \mathbb{R}^{d \times w}$ of width $w$, 
after which we add a bias and apply a nonlinearity to obtain a {\em feature map} $\mathbf{f}^k \in \mathbb{R}^{l - w + 1}$. Specifically, 
the $i$-th element of $\mathbf{f}^k$ is given by:
\begin{equation}
\mathbf{f}^k[i] = \mbox{tanh}(\langle \mathbf{C}^k[\ast, i:i+w-1] 
, \mathbf{H} \rangle + b)
\end{equation}
where $\mathbf{C}^k[\ast, i:i+w-1]$ is the $i$-to-$(i+w-1)$-th column of $\mathbf{C}^k$ and $\langle \mathbf{A}, \mathbf{B} \rangle = 
\Tr(\mathbf{A}\mathbf{B}^T)$ is the Frobenius inner product.
Finally, we take the {\em max-over-time}
\begin{equation}
y^k = \max_{i} \mathbf{f}^k[i]
\end{equation}
as the feature corresponding to the filter $\mathbf{H}$ (when applied to word $k$). The idea is to capture the most important
feature---the one with the highest value---for a given filter. A filter is essentially picking out
a character $n$-gram, where the size of the $n$-gram corresponds to the filter width.

We have described the process by which {\em one} feature is obtained from {\em one} filter matrix. Our CharCNN uses multiple filters of 
varying widths to obtain the feature vector for $k$. So if we have a total of $h$ filters $\mathbf{H}_1, \dots, \mathbf{H}_h$, 
then $\mathbf{y}^k = [y^k_1, \dots, y^k_h]$ is the input representation of $k$. For many NLP applications $h$ is typically chosen to be in $[100,1000]$.

\subsection{Highway Network}

We could simply replace $\mathbf{x}^k$ (the word embedding) with $\mathbf{y}^k$ at each $t$ in the RNN-LM, and as we show later, 
this simple model performs well on its own (Table~\ref{tab:highway}). One could also have a multilayer perceptron (MLP) over $\mathbf{y}^k$
to model interactions between the character $n$-grams picked up by the filters, but we found that this resulted in worse performance.

Instead we obtained improvements by running $\mathbf{y}^k$ through a {\em highway network}, recently proposed by 
Srivastava et al. \shortcite{Srivastava2015}. Whereas one layer of an MLP applies an affine transformation followed by a 
nonlinearity to obtain a new set of features, 
\begin{equation}
\mathbf{z} = g(\mathbf{W}\mathbf{y} + \mathbf{b})
\end{equation}
one layer of a highway network does the following:
\begin{equation}
\mathbf{z} = \mathbf{t} \odot g(\mathbf{W}_H\mathbf{y} + \mathbf{b}_H) + (\mathbf{1} -  \mathbf{t}) \odot \mathbf{y}
\end{equation}
where $g$ is a nonlinearity, $\mathbf{t} = \sigma(\mathbf{W}_T\mathbf{y} + \mathbf{b}_T)$ is called the {\em transform} gate, and 
$(\mathbf{1} -  \mathbf{t})$ is called the {\em carry} gate. Similar to the memory cells in LSTM networks, highway layers allow 
for training of deep networks by adaptively {\em carrying} some dimensions of the 
input directly to the output.\footnote{Srivastava et al. \shortcite{Srivastava2015} recommend initializing $\mathbf{b}_T$
to a negative value, in order to militate the initial behavior towards {\em carry}. We initialized $\mathbf{b}_T$ to a small interval around $-2$.} 
By construction the dimensions of $\mathbf{y}$ and $\mathbf{z}$ have to match, and 
hence $\mathbf{W}_T$ and $\mathbf{W}_H$ are square matrices. 

\section{Experimental Setup}

As is standard in language modeling, we use perplexity ($PPL$) to evaluate the performance of our models. Perplexity of a model over a sequence 
$ [w_1, \dots, w_T]$ is given by
\begin{equation}
PPL = \mbox{exp} \Big( \frac{NLL}{T} \Big)
\end{equation}
where $NLL$ is calculated over the test set. We test the model on corpora of varying languages and sizes
(statistics available in Table~\ref{tab:corpus}).

We conduct hyperparameter search, model introspection, and ablation studies on the English Penn Treebank (PTB) \cite{Marcus1993}, utilizing 
the standard training (0-20), validation (21-22), and test (23-24) splits along with pre-processing by \citeauthor{Mikolov2010} \shortcite{Mikolov2010}. 
With approximately $1$m tokens and $|\mathcal{V}|=10$k, this version has been extensively used by the language modeling 
community and is publicly available.\footnote{\url{http://www.fit.vutbr.cz/~imikolov/rnnlm/}} 

\begin{table}[!t]
\center
\tabcolsep 6.6pt
\begin{tabular}{@{}lrrcrrc@{}}
\toprule
& \multicolumn{3}{c}{\textsc{Data-s}} & \multicolumn{3}{c}{\textsc{Data-l}} \\
\addlinespace
 & $|\mathcal{V}|$ & $|\mathcal{C}|$ & $T$  & $|\mathcal{V}|$ & $|\mathcal{C}|$ & $T$ \\
\midrule
English (\textsc{En}) & $10$ k & $51$ & $1$ m & $60$ k & $197$ & $20$ m  \\ 
Czech (\textsc{Cs}) & $46$ k &  $101$ & $1$ m  & $206$ k & $195$ & $17$ m  \\ 
German (\textsc{De})  & $37$ k &  $74$ & $1$ m  & $339$ k & $260$ & $51$ m  \\ 
Spanish (\textsc{Es}) & $27$ k &  $72$ & $1$ m  & $152$ k & $222$ & $56$ m  \\ 
French (\textsc{Fr}) & $25$ k &  $76$ & $1$ m   & $137$ k & $225$ & $57$ m  \\ 
Russian (\textsc{Ru})  & $62$ k &  $62$ & $1$ m   & $497$ k & $111$ & $25$ m \\
Arabic (\textsc{Ar})  & $86$ k &  $132$ & $4$ m   & -- \hspace{2.5mm}  & -- \hspace{1mm} & --  \\
\bottomrule
\end{tabular}
\caption{Corpus statistics. $|\mathcal{V}| =$ word vocabulary size; $|\mathcal{C}| =$ character vocabulary size; $T = $ number of tokens in training set. 
The small English data is from the Penn Treebank and the Arabic data is from the News-Commentary corpus. The rest
 are from the 2013 ACL Workshop on Machine Translation. $|\mathcal{C}|$ is large because of (rarely occurring) special characters.} 
\label{tab:corpus}
\end{table}

With the optimal hyperparameters tuned on PTB, we apply the model to various morphologically rich languages: Czech, German, French, Spanish, Russian, and Arabic. 
Non-Arabic data comes from the 2013 ACL Workshop on Machine Translation,\footnote{\url{http://www.statmt.org/wmt13/translation-task.html}} and we use the
same train/validation/test splits as in \citeauthor{Botha2014} \shortcite{Botha2014}. While the raw data are publicly available, we obtained the 
preprocessed versions from the authors,\footnote{\url{http://bothameister.github.io/}} 
whose morphological NLM serves as a baseline for our work. We train on both the small datasets (\textsc{Data-s}) 
with $1$m tokens per language, and the large datasets
(\textsc{Data-l}) including the large English data which has a much bigger $|\mathcal{V}|$ than the PTB. Arabic data comes from the News-Commentary 
corpus,\footnote{\url{http://opus.lingfil.uu.se/News-Commentary.php}} and we perform our own preprocessing and train/validation/test splits.

In these datasets  only singleton words were replaced with  \texttt{\small <}\textsf{\small unk}\texttt{\small >} and hence we effectively use the full vocabulary. 
It is worth noting that the character model can utilize surface forms of OOV tokens (which were replaced with \texttt{\small <}\textsf{\small unk}\texttt{\small >}),
but we do not do this and stick to the preprocessed versions (despite disadvantaging the character models) for exact comparison against prior work.

\subsection{Optimization}
The models are trained by truncated backpropagation through time \cite{Werbos1990,Graves2013}. We backpropagate for $35$ time steps using stochastic 
gradient descent where the learning rate is initially set to $1.0$ and halved if the perplexity does not decrease by more than $1.0$ on the validation 
set after an epoch. On \textsc{Data-s} we use a batch size of $20$ and on \textsc{Data-l} we use a batch size of $100$ (for greater efficiency). 
Gradients are averaged over each batch. We train for $25$ epochs on non-Arabic and $30$ epochs on Arabic data
(which was sufficient for convergence), picking the best 
performing model on the validation set. Parameters of the model are randomly initialized over a uniform distribution with support $[-0.05, 0.05]$. 

For regularization we use dropout \cite{Hinton2012} with probability $0.5$ on the LSTM input-to-hidden layers (except on the initial Highway to LSTM 
layer) and the hidden-to-output softmax layer. We further constrain the norm of the gradients to be below $5$, so that if the $L_2$ 
norm of the gradient exceeds $5$ then we renormalize it to have  $||\cdot|| = 5$ before updating. The gradient norm constraint was crucial in training 
the model. These choices were largely guided by previous work of Zaremba et al. \shortcite{Zaremba2014} on word-level language modeling with LSTMs.

\begin{table}[!t]
\center
\begin{tabular}{llll}
\toprule
 \multicolumn{2}{c}{}  & \multicolumn{1}{c}{Small} & \multicolumn{1}{c}{Large}  \\ 
\midrule
\multirow{4}{*}{CNN} & $d$ & $15$ & $15$ \\
 & $w$ & $[1,2,3,4,5,6]$ & $[1,2,3,4,5,6,7]$ \\
 & $h$ & $[25\cdot w]$  & $[\mbox{min} \{200, 50\cdot w\}]$ \\
 & $f$ & tanh & tanh \\ 
\midrule
\multirow{2}{*}{Highway} & $l$ & 1 & 2 \\
& $g$ & ReLU & ReLU \\ 
\midrule
\multirow{2}{*}{LSTM} & $l$ & 2 & 2 \\
& $m$ & 300 & 650 \\
\bottomrule
\end{tabular}
\caption{Architecture of the small and large models. $d = $  dimensionality of character embeddings; $w =$ filter widths; $h = $ number 
of filter matrices, as a function of filter width (so the large model has filters of width $[1,2,3,4,5,6,7]$ of size $[50,100,150,200,200,200,200]$ 
for a total of $1100$ filters); $f,g = $ nonlinearity functions; $l =$ number of layers; $m = $ number of hidden units.}
\label{tab:hyper}
\end{table}

Finally, in order to speed up training on \textsc{Data-l} we employ a hierarchical softmax \cite{Morin2005}---a common strategy for training 
language models with very large $|\mathcal{V}|$---instead of the usual softmax. We pick the number of clusters $c = \lceil \sqrt{|\mathcal{V}|} \rceil$ 
and randomly split $\mathcal{V}$ into mutually exclusive and collectively exhaustive subsets $\mathcal{V}_1, \dots, \mathcal{V}_c$ of (approximately) 
equal size.\footnote{While Brown clustering/frequency-based clustering is commonly used in the literature (e.g. \citeauthor{Botha2014} \shortcite{Botha2014}
use Brown clusering), we used random clusters as our implementation 
enjoys the best speed-up when the number of words in each cluster is approximately equal. We found random clustering to work surprisingly well.} 
Then $\mbox{Pr}(w_{t+1} = j|w_{1:t})$ becomes,
\begin{equation}
\begin{split}
\mbox{Pr}(w_{t+1} = j|w_{1:t}) &= \frac{\mbox{exp}(\mathbf{h}_t \cdot \mathbf{s}^r + t^r)}{\sum_{r'=1}^c 
\mbox{exp}(\mathbf{h}_t \cdot \mathbf{s}^{r'} + t^{r'})} \\ 
&\times \frac{\mbox{exp}(\mathbf{h}_t \cdot \mathbf{p}^j_r + q^j_r)}{\sum_{j' \in \mathcal{V}_r} 
\mbox{exp}(\mathbf{h}_t \cdot \mathbf{p}^{j'}_r + q^{j'}_r)}
\end{split}
\end{equation}
where $r$ is the cluster index such that $j \in \mathcal{V}_r$. The first term is simply the probability of picking cluster $r$, and the 
second term is the probability of picking word $j$ given that cluster $r$ is picked. We found that hierarchical softmax was not 
necessary for models trained on \textsc{Data-s}.

\section{Results}

\subsection{English Penn Treebank}
\begin{table}[!t]
\center
\begin{tabular}{lrr}
\toprule
 & $PPL$ & Size \\ 
\midrule
LSTM-Word-Small & $97.6$ & $5$ m\\
LSTM-Char-Small & $92.3$ & $5$ m\\
LSTM-Word-Large & $85.4$ & $20$ m\\
LSTM-Char-Large & $78.9$ & $19$ m\\ 
\midrule
KN-$5$ (Mikolov et al. 2012) & $141.2$ & $2$ m\\
RNN$^\dagger$ (Mikolov et al. 2012)& $124.7$ & $6$ m \\
RNN-LDA$^\dagger$  (Mikolov et al. 2012) & $113.7$ & $7$ m\\
genCNN$^\dagger$ \cite{Wang2015} & $116.4$ & $8$ m \\
FOFE-FNNLM$^\dagger$  \cite{Shang2015} & $108.0$ & $6$ m\\
Deep RNN \cite{Pascanu2013} & $107.5$ & $6$ m\\
Sum-Prod Net$^\dagger$  \cite{Cheng2014} & $100.0$ & $5$ m\\
LSTM-1$^\dagger$  (Zaremba et al. 2014) & $82.7$ & $20$ m\\
LSTM-2$^\dagger$  (Zaremba et al. 2014) & $78.4$ & $52$ m \\
\bottomrule
\end{tabular}
\caption{Performance of our model versus other neural language models on the English Penn Treebank test set. 
$PPL$ refers to perplexity (lower is better) and size refers to the approximate number of parameters
in the model. KN-$5$ is a Kneser-Ney $5$-gram language model which serves as a non-neural baseline.
$^\dagger$For these models the authors did not explicitly state the number of parameters, and hence sizes shown here are estimates
based on our understanding of their papers or private correspondence with the respective authors.}
\label{tab:ptb}
\end{table}

We train two versions of our model to assess the trade-off between performance and size. Architecture of the small (LSTM-Char-Small) 
and large (LSTM-Char-Large) models is summarized in Table~\ref{tab:hyper}. As another baseline, we also train two comparable LSTM 
models that use word embeddings only (LSTM-Word-Small, LSTM-Word-Large). LSTM-Word-Small uses $200$ hidden units and 
LSTM-Word-Large uses $650$ hidden units. Word embedding sizes are also $200$ and $650$ respectively. 
These were chosen to keep the number of parameters similar to the corresponding character-level model. 

As can be seen from Table~\ref{tab:ptb}, our large model is on par with the existing state-of-the-art 
(Zaremba et al. 2014), despite having approximately $60\%$ fewer parameters. 
Our small model significantly outperforms other NLMs of similar size, even though it is penalized by the fact that 
the dataset already has OOV words replaced with \texttt{\small <}\textsf{\small unk}\texttt{\small >} (other models are purely word-level 
models).  While lower perplexities have been reported with model ensembles \cite{Mikolov2012a}, we do not include them here as
they are not comparable to the current work.

\subsection{Other Languages}
The model's performance on the English PTB is informative to the extent that it facilitates comparison against the large body of 
existing work. However, English is relatively simple from a morphological standpoint, and thus our next set of results (and arguably the 
main contribution of this paper) is focused on languages with richer morphology (Table~\ref{tab:others}, Table~\ref{tab:others2}). 

\begin{table}[!t]
\center
\tabcolsep 6pt
\begin{tabular}{@{}llcccccc@{}}
\toprule
& & \multicolumn{6}{c}{\textsc{Data-s}} \\
\addlinespace
&& \textsc{Cs} & \textsc{De} & \textsc{Es} & \textsc{Fr} &  \textsc{Ru} & \textsc{Ar} \\
\midrule
\multirow{2}{*}{Botha} & KN-$4$ & $545$ & $366$ & $241$ & $274$ & $396$ & $323$ \\
&  MLBL & $465$ & $296$ & $200$ & $225$ & $304$ & -- \\
\midrule
\multirow{3}{*}{Small}&  Word & $503$ & $305$ & $212$ & $229$ & $352$ & $216$ \\
&  Morph & $414$ & $278$ & $197$ & $216$ & $290$ & $230$ \\
&  Char & $401$ & $260$ & $182$ & $189$ & $278$ & $196$ \\
\addlinespace
\multirow{3}{*}{Large}&  Word & $493$ & $286$ & $200$ & $222$ & $357$ & $172$ \\
&  Morph & $398$ & $263$ & $177$ & $196$ & $271$ & $\mathbf{148}$ \\
&  Char &$\mathbf{371}$ & $\mathbf{239}$ &$\mathbf{165}$& $\mathbf{184}$  & $\mathbf{261}$ & $\mathbf{148}$ \\
\bottomrule
\end{tabular}
\caption{Test set perplexities for \textsc{Data-s}. First two rows are from \citeauthor{Botha2014b} \shortcite{Botha2014b} (except on Arabic 
where we trained our own KN-$4$ model) while the last six are from this paper.
KN-$4$ is a Kneser-Ney $4$-gram language model, and MLBL is the best performing morphological logbilinear model from
Botha \shortcite{Botha2014b}. Small/Large refer to model size (see Table~\ref{tab:hyper}), and 
Word/Morph/Char are models with words/morphemes/characters as inputs respectively.
}
\label{tab:others}
\end{table}

We compare our results against the morphological log-bilinear (MLBL) model from \citeauthor{Botha2014}
\shortcite{Botha2014}, whose model also takes into account subword information through 
morpheme embeddings that are summed at the input and output layers. As comparison against the MLBL models is confounded by our use of LSTMs---widely 
known to outperform their feed-forward/log-bilinear cousins---we also train an LSTM version of the morphological NLM, 
where the input representation of a word given to the LSTM is a summation of the word's morpheme embeddings.  Concretely, suppose that  
$\mathcal{M}$ is the set of morphemes in a language, $\mathbf{M} \in\mathbb{R}^{n \times |\mathcal{M}|}$ is the 
matrix of morpheme embeddings, and $\mathbf{m}^j$ is the $j$-th column of $\mathbf{M}$ (i.e. a morpheme embedding). 
Given the input word $k$, we feed the following representation to the LSTM:
\begin{equation}
\mathbf{x}^k + \sum_{j \in \mathcal{M}_k} \mathbf{m}^j
\end{equation}
where $\mathbf{x}^k$ is the word embedding (as in a word-level model) and $\mathcal{M}_k \subset \mathcal{M}$ is the set of morphemes for word $k$. The morphemes 
are obtained by running an unsupervised morphological tagger as a preprocessing step.\footnote{We use {\em Morfessor Cat-MAP} \cite{Creutz2007}, 
as in \citeauthor{Botha2014} \shortcite{Botha2014}.} We emphasize that the word embedding itself (i.e. $\mathbf{x}^k$) is added on top of the morpheme 
embeddings, as was done in  Botha and Blunsom \shortcite{Botha2014}. The morpheme embeddings are of size $200$/$650$ for the small/large models 
respectively. We further train word-level LSTM models as another baseline.

On \textsc{Data-s} it is clear from Table~\ref{tab:others} that the character-level models outperform their word-level counterparts 
despite, again, being smaller.\footnote{The difference in parameters is greater for non-PTB corpora as the size of the word model scales 
faster with $|\mathcal{V}|$. For example, on Arabic the small/large word models have $35$m/$121$m parameters while the corresponding
character models have $29$m/$69$m parameters respectively.} The character models also outperform their morphological counterparts (both MLBL and LSTM 
architectures), although improvements over the morphological LSTMs are more measured. Note that the morpheme models have strictly more parameters 
than the word models because word embeddings are used as part of the input.

Due to memory constraints\footnote{All models were trained on GPUs with 2GB memory.} we only train the small models on 
\textsc{Data-l} (Table~\ref{tab:others2}). Interestingly we do not observe significant differences going from word 
to morpheme LSTMs on Spanish, French, and English. The character models again outperform the word/morpheme models. We also 
observe significant perplexity reductions even on English when $\mathcal{V}$ is large. We conclude this section by noting that we used the same 
architecture for all languages and did not perform any language-specific tuning of hyperparameters. 

\begin{table}[!t]
\center
\tabcolsep 6pt
\begin{tabular}{@{}llcccccc@{}}
\toprule
& & \multicolumn{6}{c}{\textsc{Data-l}} \\
\addlinespace
& & \textsc{Cs} & \textsc{De} & \textsc{Es} & \textsc{Fr} &  \textsc{Ru} & \textsc{En} \\
\midrule
\multirow{2}{*}{Botha} & KN-$4$ & $862$ & $463$ & $219$ & $243$ & $390$ & $291$ \\
& MLBL & $643$ & $404$ & $203$ & $227$ & $\mathbf{300}$ & $273$ \\
\midrule
\multirow{3}{*}{Small} & Word & $701$ & $347$ & $186$ & $202$ & $353$ & $236$ \\
&  Morph & $615$ & $331$ & $189$ & $209$ & $331$ & $233$ \\
&  Char & $\mathbf{578}$ & $\mathbf{305}$ & $\mathbf{169}$ & $\mathbf{190}$ & $313$ & $\mathbf{216}$ \\
\bottomrule
\end{tabular}
\caption{Test set perplexities on \textsc{Data-l}. First two rows are from \citeauthor{Botha2014b} \shortcite{Botha2014b},
while the last three rows are from the small LSTM models described in the paper.
KN-$4$ is a Kneser-Ney $4$-gram language model, and MLBL is the best performing morphological logbilinear model from
\citeauthor{Botha2014b} \shortcite{Botha2014b}. Word/Morph/Char are models with words/morphemes/characters as inputs respectively.}
\label{tab:others2}
\end{table}

\begin{table*}[!t]
\center
\small
\em
\begin{tabular}{ccccccccc}
\toprule
& \multicolumn{5}{c}{{\em {\normalsize In Vocabulary}}} & \multicolumn{3}{c}{{\em {\normalsize Out-of-Vocabulary}}} \\
\addlinespace
& while & his & you & richard & trading & computer-aided & misinformed & looooook \\
\cmidrule(lr){2-6} \cmidrule(lr){7-9}
\multirow{4}{*}{{\em {\normalsize LSTM-Word}}} & although & your & conservatives & jonathan & advertised & -- & -- & --\\
			& letting & her & we & robert & advertising & -- & -- & -- \\
			& though & my & guys & neil & turnover & -- & -- & --\\
			& minute & their & i & nancy & turnover & -- & -- & --\\
\addlinespace
\addlinespace
	& chile & this & your & hard & heading & computer-guided & informed & look \\
{\em {\normalsize LSTM-Char}} 	& whole & hhs & young & rich& training & computerized & performed  & cook\\
{\em {\normalsize (before highway)}}	& meanwhile & is & four & richer & reading & disk-drive & transformed & looks\\
			& white & has & youth & richter & leading & computer & inform & shook \\
\addlinespace
\addlinespace
			& meanwhile & hhs & we & eduard & trade & computer-guided & informed & look\\
{\em {\normalsize LSTM-Char}}	& whole & this & your & gerard & training & computer-driven & performed & looks  \\
{\em {\normalsize (after highway)}}	& though & their & doug & edward & traded & computerized& outperformed & looked \\
			& nevertheless & your & i & carl & trader & computer & transformed & looking \\
\bottomrule
\end{tabular}
\caption{Nearest neighbor words (based on cosine similarity) of word representations from the large word-level and character-level 
(before and after highway layers) models trained on the PTB. Last three words are OOV words, 
and therefore they do not have representations in the word-level model.}
\label{tab:nn}
\end{table*}

\section{Discussion}                                  
\subsection{Learned Word Representations}

We explore the word representations learned by the models on the PTB. Table~\ref{tab:nn} has
the nearest neighbors of word representations learned from both the word-level and character-level models. For the character models 
we compare the representations obtained before and after highway layers.

Before the highway layers the representations seem to solely rely on surface forms---for example the nearest neighbors of {\em you} are 
{\em your, young, four, youth}, which are close to {\em you} in terms of edit distance. The highway layers however, seem to enable 
encoding of semantic features that are not discernable from orthography alone. After highway layers the nearest neighbor of {\em you} 
is {\em we}, which is orthographically distinct from {\em you}. Another example is {\em while} and {\em though}---these words are far 
apart edit distance-wise yet the composition model is able to place them near each other. The model also makes some clear mistakes 
(e.g. {\em his} and {\em hhs}), highlighting the limits of our approach, although this could be due to the small dataset.

The learned representations of OOV words ({\em computer-aided}, {\em misinformed}) are positioned near words with the same 
part-of-speech.  The model is also able to correct for incorrect/non-standard spelling ({\em looooook}), 
indicating potential applications for text normalization in noisy domains.

\subsection{Learned Character $N$-gram Representations}

As discussed previously, each filter of the CharCNN is essentially learning to detect particular character $n$-grams.
Our initial expectation was that each filter would learn to activate on different morphemes and then build up semantic representations 
of words from the identified morphemes. However, upon reviewing the character $n$-grams picked up by the filters
(i.e. those that maximized the value of the filter), we found that they did not (in general) correspond to valid morphemes.

\begin{figure}[!t]
\center
\includegraphics[scale=0.45]{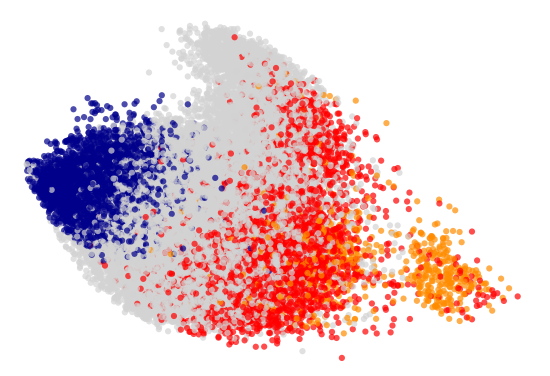}
\caption{Plot of character $n$-gram representations via PCA for English. Colors correspond to: prefixes (red), suffixes (blue), hyphenated (orange),
and all others (grey). Prefixes refer to character $n$-grams which start with the start-of-word character. Suffixes likewise refer to character
$n$-grams which end with the end-of-word character.}
\label{fig:pca}
\end{figure}

To get a better intuition for what the character composition model is learning, we plot the learned representations of 
all character $n$-grams (that occurred as part of at least two words in $\mathcal{V}$) via principal components analysis (Figure~\ref{fig:pca}). 
We feed each character $n$-gram into the CharCNN and use the CharCNN's output as the fixed dimensional
representation for the corresponding character $n$-gram. As is apparent from Figure~\ref{fig:pca}, the model 
learns to differentiate between prefixes (red), suffixes (blue), and others (grey). We also find that the representations are 
particularly sensitive to character $n$-grams containing hyphens (orange), presumably because this is a strong signal of a word's part-of-speech.

\subsection{Highway Layers}
We quantitatively investigate the effect of highway network layers via ablation studies (Table~\ref{tab:highway}). We train a model without 
any highway layers, and find that performance decreases significantly. As the difference in performance could be due to the decrease in model size, 
we also train a model that feeds $\mathbf{y}^k$ (i.e. word representation from the CharCNN) through a one-layer multilayer perceptron (MLP) to use 
as input into the LSTM. We find that the MLP does poorly, although this could be due to optimization issues.

We hypothesize that highway networks are especially well-suited to work with CNNs, 
adaptively combining local features detected by the individual filters. CNNs have already proven to be been successful for many NLP tasks 
\cite{Collobert2011,Shen2014,Kalchbrenner2014,Kim2014,Zhang2015,Lei2015}, and we posit 
that further gains could be achieved by employing highway layers on top of existing CNN architectures.

We also anecdotally note that (1) having one to two highway layers was important, but more highway layers generally resulted in similar performance 
(though this may depend on the size of the datasets), (2) having more convolutional layers before max-pooling did not help, and (3) highway layers 
did not improve models that only used word embeddings as inputs. 

\begin{table}[!t]
\center
\begin{tabular}{lrc}
\toprule
& \multicolumn{2}{c}{LSTM-Char} \\
\addlinespace
 & Small & Large \\ 
\midrule
No Highway Layers & $100.3$ & $84.6$ \\
One Highway Layer & $92.3$ & $79.7$\\
Two Highway Layers & $90.1$ & $78.9$\\
One MLP Layer & $111.2$ &$92.6$ \\ 
\bottomrule
\end{tabular}
\caption{Perplexity on the Penn Treebank for small/large models trained with/without highway layers.}
\label{tab:highway}
\end{table}

\subsection{Effect of Corpus/Vocab Sizes}
We next study the effect of training corpus/vocabulary sizes on the relative performance 
between the different models. We take the German (\textsc{De}) dataset from \textsc{Data-l} and vary the training corpus/vocabulary sizes, calculating
the perplexity reductions as a result of going from a small word-level model to a small character-level model. To vary the vocabulary size 
we take the most frequent $k$ words and replace the rest with \texttt{\small <}\textsf{\small unk}\texttt{\small >}. As with previous experiments
the character model does not utilize surface forms of \texttt{\small <}\textsf{\small unk}\texttt{\small >} and simply treats it as another token.
Although Table~\ref{tab:relperf} 
suggests that the perplexity reductions become less pronounced as the corpus size increases, we nonetheless find that the character-level 
model outperforms the word-level model in all scenarios.

\begin{table}[!t]
\center
\begin{tabular}{crrrrc}
\toprule
  && \multicolumn{4}{c}{$|\mathcal{V}|$} \\
\addlinespace
 &&  $10$ k & $25$ k &$50$ k & $100$ k \\ 
\midrule
\multirow{4}{*}{$T$} &$1$ m & $17\%$ & $16\%$ &$21\%$ & --\\
&$5$ m & $8\%$ & $14\%$ & $16\%$ & $21\%$\\
&$10$ m & $9\%$ & $9\%$ & $12\%$&  $15\%$\\
&$25$ m & $9\%$ & $8\%$ &  $9\%$&  $10\%$\\
\bottomrule
\end{tabular}
\caption{Perplexity reductions by going from small word-level to character-level models based on different corpus/vocabulary sizes on German 
(\textsc{De}). $|\mathcal{V}|$ is the vocabulary size and $T$ is the number of tokens in the training set. The full vocabulary of the $1$m dataset
was less than $100$k and hence that scenario is unavailable.}
\label{tab:relperf}
\end{table}

\subsection{Further Observations}
We report on some further experiments and observations:
\begin{itemize}
\item Combining word embeddings with the CharCNN's output to form a combined representation of a word (to be used as input to the LSTM) 
resulted in slightly worse performance ($81$ on PTB with a large model). This was surprising, as improvements 
have been reported on part-of-speech tagging \cite{Santos2014a}
and named entity recognition \cite{Santos2015} by concatenating word embeddings with the output from a character-level CNN. While this could be 
due to insufficient experimentation on our part,\footnote{We experimented with (1) concatenation, (2) tensor
products, (3) averaging, and (4) adaptive weighting schemes whereby the model learns a convex combination of word embeddings 
and the CharCNN outputs.} it suggests that for some tasks, word embeddings are superfluous---character inputs are good enough. 

\item While our model requires additional convolution operations over characters and is thus 
slower than a comparable word-level model which can perform a simple lookup at the input layer, we found that the difference was manageable
with optimized GPU implementations---for example on PTB the large character-level model trained at $1500$ tokens/sec compared to the
word-level model which trained at $3000$ tokens/sec. For scoring, our model can have the same running time as a pure word-level model, as 
the CharCNN's outputs can be pre-computed for all words in $\mathcal{V}$. This would, however, be at the expense of increased model size, 
and thus a trade-off can be made between run-time speed and memory (e.g. one could restrict the pre-computation to the most frequent words).
\end{itemize}

\section{Related Work}

Neural Language Models (NLM) encompass a rich family  of neural network architectures for language modeling. Some example 
architectures include feed-forward \cite{Bengio2003}, recurrent \cite{Mikolov2010}, sum-product \cite{Cheng2014}, log-bilinear 
\cite{Mnih2007}, and convolutional \cite{Wang2015} networks. 

In order to address the rare word problem, \citeauthor{Alexandrescu2006} \shortcite{Alexandrescu2006}---building on analogous work 
on count-based $n$-gram language models by Bilmes and Kirchhoff \shortcite{Bilmes2003}---represent a word as a set of 
shared factor embeddings. Their Factored Neural Language Model (FNLM) can incorporate morphemes, word shape information 
(e.g. capitalization) or any other annotation (e.g. part-of-speech tags) to represent words. 

A specific class of FNLMs leverages morphemic information by viewing a word as a function of its (learned) morpheme embeddings
\cite{Luong2013,Botha2014,Qui2014}. For example \citeauthor{Luong2013} \shortcite{Luong2013} apply a recursive neural 
network over morpheme embeddings to obtain the embedding for a single word. While such models have proved useful, they require morphological 
tagging as a preprocessing step.

Another direction of work has involved purely character-level NLMs, wherein both input and output are 
characters \cite{Sutskever2011,Graves2013}. Character-level models obviate the need for morphological tagging
or manual feature engineering, and have the attractive property of being able to generate novel words.
However they are generally outperformed by word-level models \cite{Mikolov2012b}.

Outside of language modeling, improvements have been reported on part-of-speech tagging \cite{Santos2014a} and
named entity recognition \cite{Santos2015} by representing a word as a concatenation of its word embedding and an output 
from a character-level CNN, and using the combined representation as features in a 
Conditional Random Field (CRF). \citeauthor{Zhang2015} \shortcite{Zhang2015} do away with word embeddings completely and 
show that for text classification, a deep CNN over characters performs well. \citeauthor{Ballesteros2015} \shortcite{Ballesteros2015}
use an RNN over characters only to train a transition-based parser, obtaining improvements on many morphologically rich languages. 

Finally, \citeauthor{Ling2015} \shortcite{Ling2015} apply a bi-directional LSTM over characters to use as inputs for language modeling 
and part-of-speech tagging. They show improvements on various languages (English, Portuguese, Catalan, German, Turkish).
It remains open as to which character composition model (i.e. CNN or LSTM) performs better.

\section{Conclusion}
We have introduced a neural language model that utilizes only character-level inputs. Predictions are still made at the word-level. 
Despite having fewer parameters, our model outperforms baseline models that utilize word/morpheme embeddings in the input layer.
Our work questions the necessity of word embeddings (as inputs) for neural language modeling.

Analysis of word representations obtained from the character composition part of the model further indicates that the model 
is able to encode, from characters only, rich semantic and orthographic features. Using the CharCNN and highway layers for 
representation learning (e.g. as input into \textsf{\small word2vec} \cite{Mikolov2013a}) remains an avenue for future work.
 
Insofar as sequential processing of words as inputs is ubiquitous in natural language processing, 
it would be interesting to see if the architecture introduced in this paper is viable for other tasks---for example, as an encoder/decoder 
in neural machine translation \cite{Cho2014,Sutskever2014}.

\section*{Acknowledgments}

We are especially grateful to Jan Botha for providing the preprocessed datasets and the model results.

\bibliographystyle{aaai}
\small
\bibliography{master}

\begin{thebibliography}{}

\bibitem[\protect\citeauthoryear{Alexandrescu and
  Kirchhoff}{2006}]{Alexandrescu2006}
Alexandrescu, A., and Kirchhoff, K.
\newblock 2006.
\newblock {F}actored {N}eural {L}anguage {M}odels.
\newblock In {\em Proceedings of NAACL}.

\bibitem[\protect\citeauthoryear{Ballesteros, Dyer, and
  Smith}{2015}]{Ballesteros2015}
Ballesteros, M.; Dyer, C.; and Smith, N.~A.
\newblock 2015.
\newblock {I}mproved {T}ransition-{B}ased {P}arsing by {M}odeling {C}haracters
  instead of {W}ords with {LSTM}s.
\newblock In {\em Proceedings of EMNLP}.

\bibitem[\protect\citeauthoryear{Bengio, Ducharme, and
  Vincent}{2003}]{Bengio2003}
Bengio, Y.; Ducharme, R.; and Vincent, P.
\newblock 2003.
\newblock {A} {N}eural {P}robabilistic {L}anguage {M}odel.
\newblock {\em Journal of Machine Learning Research} 3:1137--1155.

\bibitem[\protect\citeauthoryear{Bengio, Simard, and
  Frasconi}{1994}]{Bengio1994}
Bengio, Y.; Simard, P.; and Frasconi, P.
\newblock 1994.
\newblock {L}earning {L}ong-term {D}ependencies with {G}radient {D}escent is
  {D}ifficult.
\newblock {\em IEEE Transactions on Neural Networks} 5:157--166.

\bibitem[\protect\citeauthoryear{Bilmes and Kirchhoff}{2003}]{Bilmes2003}
Bilmes, J., and Kirchhoff, K.
\newblock 2003.
\newblock {F}actored {L}anguage {M}odels and {G}eneralized {P}arallel
  {B}ackoff.
\newblock In {\em Proceedings of NAACL}.

\bibitem[\protect\citeauthoryear{Botha and Blunsom}{2014}]{Botha2014}
Botha, J., and Blunsom, P.
\newblock 2014.
\newblock {C}ompositional {M}orphology for {W}ord {R}epresentations and
  {L}anguage {M}odelling.
\newblock In {\em Proceedings of ICML}.

\bibitem[\protect\citeauthoryear{Botha}{2014}]{Botha2014b}
Botha, J.
\newblock 2014.
\newblock {P}robabilistic {M}odelling of {M}orphologically {R}ich {L}anguages.
\newblock {\em DPhil Dissertation, Oxford University}.

\bibitem[\protect\citeauthoryear{Chen and Goodman}{1998}]{Chen1998}
Chen, S., and Goodman, J.
\newblock 1998.
\newblock {A}n {E}mpirical {S}tudy of {S}moothing {T}echniques for {L}anguage
  {M}odeling.
\newblock {\em Technical Report, Harvard University}.

\bibitem[\protect\citeauthoryear{Cheng \bgroup et al\mbox.\egroup
  }{2014}]{Cheng2014}
Cheng, W.~C.; Kok, S.; Pham, H.~V.; Chieu, H.~L.; and Chai, K.~M.
\newblock 2014.
\newblock {L}anguage {M}odeling with {S}um-{P}roduct {N}etworks.
\newblock In {\em Proceedings of INTERSPEECH}.

\bibitem[\protect\citeauthoryear{Cho \bgroup et al\mbox.\egroup
  }{2014}]{Cho2014}
Cho, K.; van Merrienboer, B.; Gulcehre, C.; Bahdanau, D.; Bougares, F.;
  Schwenk, H.; and Bengio, Y.
\newblock 2014.
\newblock {L}earning {P}hrase {R}epresentations using {RNN} {E}ncoder-{D}ecoder
  for {S}tatistical {M}achine {T}ranslation.
\newblock In {\em Proceedings of EMNLP}.

\bibitem[\protect\citeauthoryear{Collobert \bgroup et al\mbox.\egroup
  }{2011}]{Collobert2011}
Collobert, R.; Weston, J.; Bottou, L.; Karlen, M.; Kavukcuoglu, K.; and Kuksa,
  P.
\newblock 2011.
\newblock {N}atural {L}anguage {P}rocessing (almost) from {S}cratch.
\newblock {\em Journal of Machine Learning Research} 12:2493--2537.

\bibitem[\protect\citeauthoryear{Creutz and Lagus}{2007}]{Creutz2007}
Creutz, M., and Lagus, K.
\newblock 2007.
\newblock {U}nsupervised {M}odels for {M}orpheme {S}egmentation and
  {M}orphology {L}earning.
\newblock In {\em Proceedings of the ACM Transations on Speech and Language
  Processing}.

\bibitem[\protect\citeauthoryear{Deerwester, Dumais, and
  Harshman}{1990}]{Deerwester1990}
Deerwester, S.; Dumais, S.; and Harshman, R.
\newblock 1990.
\newblock {I}ndexing by {L}atent {S}emantic {A}nalysis.
\newblock {\em Journal of American Society of Information Science} 41:391--407.

\bibitem[\protect\citeauthoryear{dos Santos and Guimaraes}{2015}]{Santos2015}
dos Santos, C.~N., and Guimaraes, V.
\newblock 2015.
\newblock {B}oosting {N}amed {E}ntity {R}ecognition with {N}eural {C}haracter
  {E}mbeddings.
\newblock In {\em Proceedings of ACL Named Entities Workshop}.

\bibitem[\protect\citeauthoryear{dos Santos and Zadrozny}{2014}]{Santos2014a}
dos Santos, C.~N., and Zadrozny, B.
\newblock 2014.
\newblock {L}earning {C}haracter-level {R}epresentations for {P}art-of-{S}peech
  {T}agging.
\newblock In {\em Proceedings of ICML}.

\bibitem[\protect\citeauthoryear{Graves}{2013}]{Graves2013}
Graves, A.
\newblock 2013.
\newblock {G}enerating {S}equences with {R}ecurrent {N}eural {N}etworks.
\newblock {\em arXiv:1308.0850}.

\bibitem[\protect\citeauthoryear{Hinton \bgroup et al\mbox.\egroup
  }{2012}]{Hinton2012}
Hinton, G.; Srivastava, N.; Krizhevsky, A.; Sutskever, I.; and Salakhutdinov,
  R.
\newblock 2012.
\newblock {I}mproving {N}eural {N}etworks by {P}reventing {C}o-{A}daptation of
  {F}eature {D}etectors.
\newblock {\em arxiv:1207.0580}.

\bibitem[\protect\citeauthoryear{Hochreiter and
  Schmidhuber}{1997}]{Hochreiter1997}
Hochreiter, S., and Schmidhuber, J.
\newblock 1997.
\newblock {L}ong {S}hort-{T}erm {M}emory.
\newblock {\em Neural Computation} 9:1735--1780.

\bibitem[\protect\citeauthoryear{Kalchbrenner, Grefenstette, and
  Blunsom}{2014}]{Kalchbrenner2014}
Kalchbrenner, N.; Grefenstette, E.; and Blunsom, P.
\newblock 2014.
\newblock {A} {C}onvolutional {N}eural {N}etwork for {M}odelling {S}entences.
\newblock In {\em Proceedings of ACL}.

\bibitem[\protect\citeauthoryear{Kim}{2014}]{Kim2014}
Kim, Y.
\newblock 2014.
\newblock {C}onvolutional {N}eural {N}etworks for {S}entence {C}lassification.
\newblock In {\em Proceedings of EMNLP}.

\bibitem[\protect\citeauthoryear{Krizhevsky, Sutskever, and
  Hinton}{2012}]{Krizhevsky2012}
Krizhevsky, A.; Sutskever, I.; and Hinton, G.
\newblock 2012.
\newblock {I}mage{N}et {C}lassification with {D}eep {C}onvolutional {N}eural
  {N}etworks.
\newblock In {\em Proceedings of NIPS}.

\bibitem[\protect\citeauthoryear{LeCun \bgroup et al\mbox.\egroup
  }{1989}]{LeCun1989}
LeCun, Y.; Boser, B.; Denker, J.~S.; Henderson, D.; Howard, R.~E.; Hubbard, W.;
  and Jackel, L.~D.
\newblock 1989.
\newblock {H}andwritten {D}igit {R}ecognition with a {B}ackpropagation
  {N}etwork.
\newblock In {\em Proceedings of NIPS}.

\bibitem[\protect\citeauthoryear{Lei, Barzilay, and Jaakola}{2015}]{Lei2015}
Lei, T.; Barzilay, R.; and Jaakola, T.
\newblock 2015.
\newblock {M}olding {CNN}s for {T}ext: {N}on-linear, {N}on-consecutive
  {C}onvolutions.
\newblock In {\em Proceedings of EMNLP}.

\bibitem[\protect\citeauthoryear{Ling \bgroup et al\mbox.\egroup
  }{2015}]{Ling2015}
Ling, W.; Lui, T.; Marujo, L.; Astudillo, R.~F.; Amir, S.; Dyer, C.; Black,
  A.~W.; and Trancoso, I.
\newblock 2015.
\newblock {F}inding {F}unction in {F}orm: {C}ompositional {C}haracter {M}odels
  for {O}pen {V}ocabulary {W}ord {R}epresentation.
\newblock In {\em Proceedings of EMNLP}.

\bibitem[\protect\citeauthoryear{Luong, Socher, and Manning}{2013}]{Luong2013}
Luong, M.-T.; Socher, R.; and Manning, C.
\newblock 2013.
\newblock {B}etter {W}ord {R}epresentations with {R}ecursive {N}eural
  {N}etworks for {M}orphology.
\newblock In {\em Proceedings of CoNLL}.

\bibitem[\protect\citeauthoryear{Marcus, Santorini, and
  Marcinkiewicz}{1993}]{Marcus1993}
Marcus, M.; Santorini, B.; and Marcinkiewicz, M.
\newblock 1993.
\newblock {B}uilding a {L}arge {A}nnotated {C}orpus of {E}nglish: the {P}enn
  {T}reebank.
\newblock {\em Computational Linguistics} 19:331--330.

\bibitem[\protect\citeauthoryear{Mikolov and Zweig}{2012}]{Mikolov2012a}
Mikolov, T., and Zweig, G.
\newblock 2012.
\newblock {C}ontext {D}ependent {R}ecurrent {N}eural {N}etwork {L}anguage
  {M}odel.
\newblock In {\em Proceedings of SLT}.

\bibitem[\protect\citeauthoryear{Mikolov \bgroup et al\mbox.\egroup
  }{2010}]{Mikolov2010}
Mikolov, T.; Karafiat, M.; Burget, L.; Cernocky, J.; and Khudanpur, S.
\newblock 2010.
\newblock {R}ecurrent {N}eural {N}etwork {B}ased {L}anguage {M}odel.
\newblock In {\em Proceedings of INTERSPEECH}.

\bibitem[\protect\citeauthoryear{Mikolov \bgroup et al\mbox.\egroup
  }{2011}]{Mikolov2011}
Mikolov, T.; Deoras, A.; Kombrink, S.; Burget, L.; and Cernocky, J.
\newblock 2011.
\newblock {E}mpirical {E}valuation and {C}ombination of {A}dvanced {L}anguage
  {M}odeling {T}echniques.
\newblock In {\em Proceedings of INTERSPEECH}.

\bibitem[\protect\citeauthoryear{Mikolov \bgroup et al\mbox.\egroup
  }{2012}]{Mikolov2012b}
Mikolov, T.; Sutskever, I.; Deoras, A.; Le, H.-S.; Kombrink, S.; and Cernocky,
  J.
\newblock 2012.
\newblock {S}ubword {L}anguage {M}odeling with {N}eural {N}etworks.
\newblock {\em preprint: www.fit.vutbr.cz/\~imikolov/rnnlm/char.pdf}.

\bibitem[\protect\citeauthoryear{Mikolov \bgroup et al\mbox.\egroup
  }{2013}]{Mikolov2013a}
Mikolov, T.; Chen, K.; Corrado, G.; and Dean, J.
\newblock 2013.
\newblock {E}fficient {E}stimation of {W}ord {R}epresentations in {V}ector
  {S}pace.
\newblock {\em arXiv:1301.3781}.

\bibitem[\protect\citeauthoryear{Mnih and Hinton}{2007}]{Mnih2007}
Mnih, A., and Hinton, G.
\newblock 2007.
\newblock {T}hree {N}ew {G}raphical {M}odels for {S}tatistical {L}anguage
  {M}odelling.
\newblock In {\em Proceedings of ICML}.

\bibitem[\protect\citeauthoryear{Morin and Bengio}{2005}]{Morin2005}
Morin, F., and Bengio, Y.
\newblock 2005.
\newblock {H}ierarchical {P}robabilistic {N}eural {N}etwork {L}anguage {M}odel.
\newblock In {\em Proceedings of AISTATS}.

\bibitem[\protect\citeauthoryear{Pascanu \bgroup et al\mbox.\egroup
  }{2013}]{Pascanu2013}
Pascanu, R.; Culcehre, C.; Cho, K.; and Bengio, Y.
\newblock 2013.
\newblock {H}ow to {C}onstruct {D}eep {N}eural {N}etworks.
\newblock {\em arXiv:1312.6026}.

\bibitem[\protect\citeauthoryear{Qui \bgroup et al\mbox.\egroup
  }{2014}]{Qui2014}
Qui, S.; Cui, Q.; Bian, J.; and Gao, B.
\newblock 2014.
\newblock {C}o-learning of {W}ord {R}epresentations and {M}orpheme
  {R}epresentations.
\newblock In {\em Proceedings of COLING}.

\bibitem[\protect\citeauthoryear{Shen \bgroup et al\mbox.\egroup
  }{2014}]{Shen2014}
Shen, Y.; He, X.; Gao, J.; Deng, L.; and Mesnil, G.
\newblock 2014.
\newblock {A} {L}atent {S}emantic {M}odel with {C}onvolutional-pooling
  {S}tructure for {I}nformation {R}etrieval.
\newblock In {\em Proceedings of CIKM}.

\bibitem[\protect\citeauthoryear{Srivastava, Greff, and
  Schmidhuber}{2015}]{Srivastava2015}
Srivastava, R.~K.; Greff, K.; and Schmidhuber, J.
\newblock 2015.
\newblock {T}raining {V}ery {D}eep {N}etworks.
\newblock {\em arXiv:1507.06228}.

\bibitem[\protect\citeauthoryear{Sundermeyer, Schluter, and
  Ney}{2012}]{Sundermeyer2012}
Sundermeyer, M.; Schluter, R.; and Ney, H.
\newblock 2012.
\newblock {LSTM} {N}eural {N}etworks for {L}anguage {M}odeling.

\bibitem[\protect\citeauthoryear{Sutskever, Martens, and
  Hinton}{2011}]{Sutskever2011}
Sutskever, I.; Martens, J.; and Hinton, G.
\newblock 2011.
\newblock {G}enerating {T}ext with {R}ecurrent {N}eural {N}etworks.

\bibitem[\protect\citeauthoryear{Sutskever, Vinyals, and
  Le}{2014}]{Sutskever2014}
Sutskever, I.; Vinyals, O.; and Le, Q.
\newblock 2014.
\newblock {S}equence to {S}equence {L}earning with {N}eural {N}etworks.

\bibitem[\protect\citeauthoryear{Wang \bgroup et al\mbox.\egroup
  }{2015}]{Wang2015}
Wang, M.; Lu, Z.; Li, H.; Jiang, W.; and Liu, Q.
\newblock 2015.
\newblock $gen${CNN}: {A} {C}onvolutional {A}rchitecture for {W}ord {S}equence
  {P}rediction.
\newblock In {\em Proceedings of ACL}.

\bibitem[\protect\citeauthoryear{Werbos}{1990}]{Werbos1990}
Werbos, P.
\newblock 1990.
\newblock Back-propagation {T}hrough {T}ime: what it does and how to do it.
\newblock In {\em Proceedings of IEEE}.

\bibitem[\protect\citeauthoryear{Zaremba, Sutskever, and
  Vinyals}{2014}]{Zaremba2014}
Zaremba, W.; Sutskever, I.; and Vinyals, O.
\newblock 2014.
\newblock {R}ecurrent {N}eural {N}etwork {R}egularization.
\newblock {\em arXiv:1409.2329}.

\bibitem[\protect\citeauthoryear{Zhang \bgroup et al\mbox.\egroup
  }{2015}]{Shang2015}
Zhang, S.; Jiang, H.; Xu, M.; Hou, J.; and Dai, L.
\newblock 2015.
\newblock {T}he {F}ixed-{S}ize {O}rdinally-{F}orgetting {E}ncoding {M}ethod for
  {N}eural {N}etwork {L}anguage {M}odels.
\newblock In {\em Proceedings of ACL}.

\bibitem[\protect\citeauthoryear{Zhang, Zhao, and LeCun}{2015}]{Zhang2015}
Zhang, X.; Zhao, J.; and LeCun, Y.
\newblock 2015.
\newblock {C}haracter-level {C}onvolutional {N}etworks for {T}ext
  {C}lassification.
\newblock In {\em Proceedings of NIPS}.

\end{thebibliography}
\end{document}